\documentclass[conference]{IEEEtran}
\IEEEoverridecommandlockouts
\usepackage{cite}
\usepackage{hyperref}
\usepackage{amsmath,amssymb,amsfonts}
\usepackage{algorithmic}
\usepackage{graphicx}
\usepackage{multirow}
\usepackage{textcomp}
\usepackage{xcolor}
\usepackage{svg}
\def\BibTeX{{\rm B\kern-.05em{\sc i\kern-.025em b}\kern-.08em
    T\kern-.1667em\lower.7ex\hbox{E}\kern-.125emX}}

\title{L3D-Pose: Lifting Pose for 3D Avatars from a Single Camera in the Wild}

\author{
\IEEEauthorblockN{Soumyaratna Debnath$^*$\thanks{$^*$ These authors contributed equally to this work.}}
\IEEEauthorblockA{
\textit{IIT Gandhinagar, India}\\
debnathsoumyaratna@iitgn.ac.in}
\and
\IEEEauthorblockN{Harish Katti$^*$}
\IEEEauthorblockA{
\textit{NIMH, NIH, USA}\\ 
kattih2@nih.gov}
\and
\IEEEauthorblockN{Shashikant Verma}
\IEEEauthorblockA{
\textit{IIT Gandhinagar, India}\\
shashikant.verma@iitgn.ac.in}
\and
\IEEEauthorblockN{Shanmuganathan Raman}
\IEEEauthorblockA{
\textit{IIT Gandhinagar, India}\\
shanmuga@iitgn.ac.in}
}

\begin{document}

\maketitle

\begin{abstract}
While 2D pose estimation has advanced our ability to interpret body movements in animals and primates, it is limited by the lack of depth information, constraining its application range. 3D pose estimation provides a more comprehensive solution by incorporating spatial depth, yet creating extensive 3D pose datasets for animals is challenging due to their dynamic and unpredictable behaviours in natural settings. To address this, we propose a hybrid approach that utilizes rigged avatars and the pipeline to generate synthetic datasets to acquire the necessary 3D annotations for training. Our method introduces a simple attention-based MLP network for converting 2D poses to 3D, designed to be independent of the input image to ensure scalability for poses in natural environments. Additionally, we identify that existing anatomical keypoint detectors are insufficient for accurate pose retargeting onto arbitrary avatars. To overcome this, we present a lookup table based on a deep pose estimation method using a synthetic collection of diverse actions rigged avatars perform. Our experiments demonstrate the effectiveness and efficiency of this lookup table-based retargeting approach.  Overall, we propose a comprehensive framework with systematically synthesized datasets for lifting poses from 2D to 3D and then utilize this to re-target motion from wild settings onto arbitrary avatars. The L3d-Pose dataset can be found at \textit{{\color{magenta} \href{https://soumyaratnadebnath.github.io/L3D-Pose/}{https://soumyaratnadebnath.github.io/L3D-Pose}}}.
\end{abstract}

\section{Introduction}

The inherent limitation of 2D pose estimation \cite{jiang2023rtmpose, xu2022vitpose, xu2023vitposepp} is its inability to represent depth, which is critical for accurately understanding the three-dimensional structure of poses in real-world scenarios. As a result, relying solely on 2D coordinates often leads to ambiguities, particularly in occluded or complex body configurations, where similar 2D projections can correspond to entirely different 3D poses. To address these challenges, 3D pose estimation \cite{sun2018integral, karashchuk2021anipose} offers a more comprehensive solution by incorporating depth information, enabling a richer and more accurate understanding of movement and biological action. 3D pose estimation is essential for advancing applications in virtual reality, where precise spatial awareness is crucial.

Curating 3D pose data for animals and primates is challenging due to their unpredictable movements in natural environments. Unlike humans, animals exhibit dynamic, non-repetitive behaviours, and species diversity adds complexity to capturing accurate annotations. Traditional motion capture systems are often impractical, and camera-based methods struggle with occlusion, erratic motion, and limited viewpoints, making large-scale, high-quality 3D pose data in natural settings difficult to obtain.

Synthetic 3D data provides a valuable solution for 3D pose estimation. By leveraging synthetic environments, models can be trained using abundant 3D data. However, while efficient and diverse, synthetic data may not fully capture real-world dynamics, leading to challenges in generalizing to real-world 3D poses due to differences in appearance and lighting.

In our approach, we first train models on naturally available 2D datasets to predict accurate 2D poses. We then use the priors from the synthetically generated 3D pose data to "lift" these 2D poses into the 3D space, effectively bridging the gap between 2D and 3D pose estimation in the wild. This method allows us to leverage the abundance of open-source 2D pose data while incorporating depth and spatial insights from synthetic 3D datasets. As a result, we can achieve accurate 3D pose predictions without relying on real-world 3D annotations.

\noindent \textbf{Contributions.} In summary, our main contributions are: \textbf{1.} We introduce a complete pipeline to create synthetic datasets encompassing diverse action sets utilizing a physics-based game engine. Following this unified approach, we develop two datasets, namely, deep macaque and deep horse. \textbf{2.} We propose a lightweight Attention MLP network to lift 2D poses into 3D space independent of image data, ensuring scalability to poses occurring in the wild. Furthermore, we propose a retargeting method to transfer pose and textures onto avatars, enabling seamless graphics applications.

\begin{figure*}[t]
\centering
\includegraphics[width=0.8\textwidth]{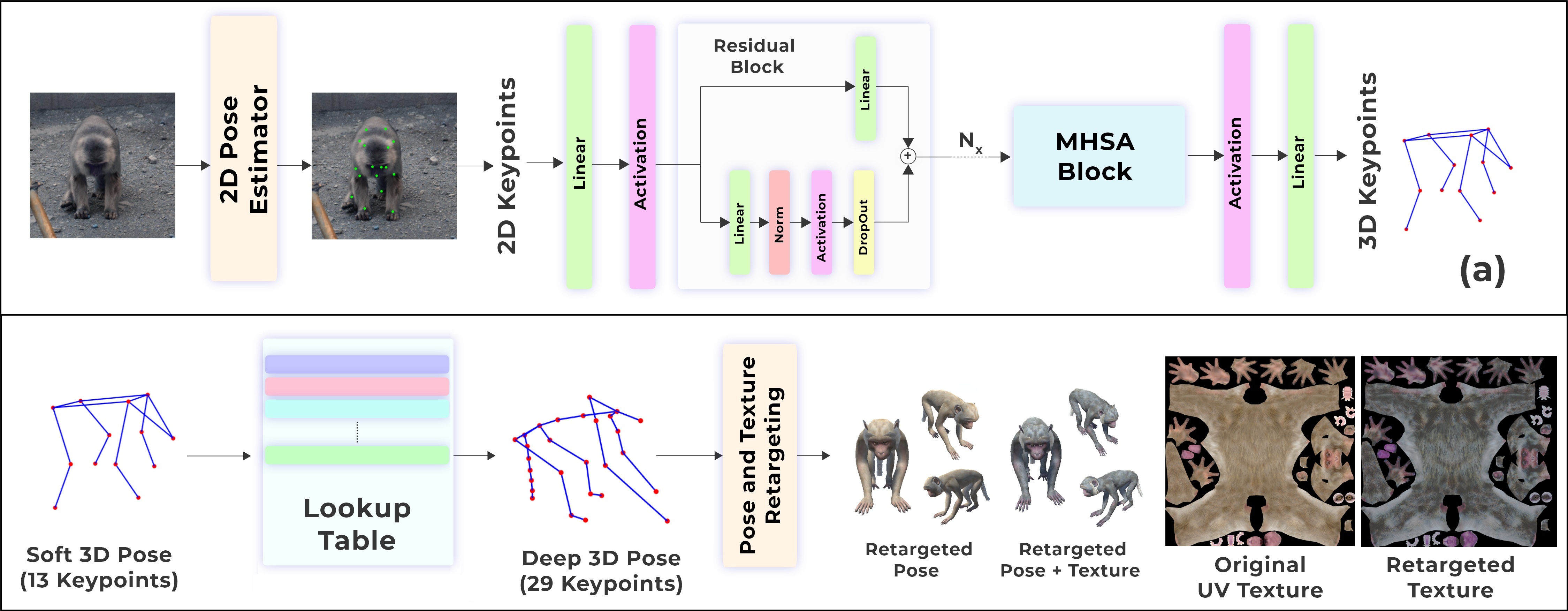}
\caption{\textbf{Process flow of the proposed methodology.} For a given natural image, we first use pre-trained 2D pose estimation techniques to obtain 2D keypoints in the image. Our attention-based simple MLP architecture, trained on a synthetic dataset, effectively lifts these normalised 2D keypoints into a partial soft 3D pose, as illustrated in (a). We then match this partial 3D pose to the closest deep pose from a look-up table, which includes a diverse set of 3D poses derived from synthetic motion sequences. The Deep 3D Pose provides the necessary information to transfer the pose from the image onto an avatar model, as demonstrated in (b). 
% The final result can be seamlessly applied to various graphics applications and behaviour analysis. 
Zooming in is recommended for better clarity.}
\label{fig:flow}
\vspace{-0.3cm}
\end{figure*}

\section{Related Work}
\noindent \textbf{3D Pose Estimation.} 
With advancements in motion capture techniques, such as those used by Vicon, Xsens, and OptiTrack, it has become easier to develop data-driven supervised methods for directly regressing 3D keypoints from images. Several methods \cite{moon2019camera, sun2018integral, zhou2019hemlets, jiang2023rtmpose} leverage deep neural networks, such as CNNs and Transformers, to estimate 3D pose. 
Most research in 3D pose estimation has concentrated on humans, as acquiring 3D data from human subjects is relatively straightforward with established protocols. However, studies on 3D animal pose estimation are limited due to the scarcity of 3D annotations for animals, which is challenging to obtain compared to humans. 
For animals, multi-perspective methods are commonly used, which involve multi-camera setups for triangulation and the use of calibration boards \cite{nath2019using, gunel2019deepfly3d, karashchuk2021anipose, bala2020openmonkeystudio, martini2024macaction}. 
However, these highly data-driven methods require expensive 3D motion acquisition setups, making them difficult to scale to other animal species.

\noindent \textbf{Lifting 2D Pose to 3D Pose.} 
Various approaches initially estimate 2D keypoints using off-the-shelf pose estimation techniques \cite{mathis2018deeplabcut,lauer2022multi,pereira2022sleap,xu2022vitpose,xu2023vitposepp} and subsequently apply methods to predict the 3D pose. 
The foundational work \cite{lee1985determination} established the basis for various subsequent studies \cite{jiang20103d,gupta20143d,chen20173d,ionescu2014iterated,ci2019optimizing} that address the inverse projection problem to estimate 3D locations from 2D keypoints. More recently, deep learning approaches have been utilized to address this problem \cite{pavlakos2017coarse,moreno20173d,martinez2017simple,pavllo20193d,wang2020motion, cai2019exploiting, xie2024hogformer,li2024graphmlp, chang20182d, hardy2024links, 10029937}, employing spatiotemporal convolutions and transformer-based methods and conditional random fields. 
While these methods have been widely used for human 3D pose analysis, their application to animal 3D pose estimation is still relatively unexplored. Our proposed approach generates 3D keypoints solely from 2D keypoints without relying on the input image. The 3D lifting is supervised using synthetically generated data, which is otherwise challenging to obtain. Leveraging anatomical constraints on pose variations and image-independent training, our method achieves realistic 3D reconstructions of 2D keypoints estimated from in-wild images.

\begin{figure*}[t]
\label{fig:results}
\centering
\includegraphics[width=0.9\textwidth]{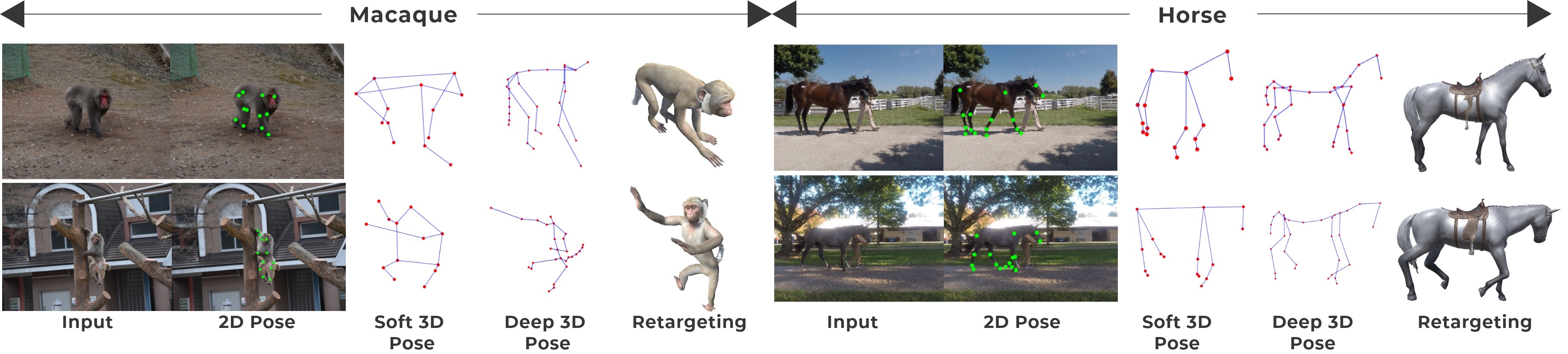}
\caption{ Lifting and retargeting of 3D Pose on rigged models obtained by the proposed framework. We present results on two assets: Macaque and Horse.
}
\label{fig:qual}
\vspace{-0.3cm}
\end{figure*}

\begin{table}[]
\centering
\caption{Index correspondence between the MacaquePose and Horse Datasets with Proposed Synthetic Datasets}
\label{tab:data-corr}
\Huge
\resizebox{0.9\columnwidth}{!}{%
\begin{tabular}{lll|lll}
\hline
\multicolumn{3}{c|}{\textbf{Macaque}}               & \multicolumn{3}{c}{\textbf{Horse}}                    \\ \hline
\multicolumn{1}{c}{\textbf{Full Macaque}} &
  \multicolumn{1}{c}{\textbf{\begin{tabular}[c]{@{}c@{}}Deep Macaque \\      (Ours)\end{tabular}}} &
  \multicolumn{1}{c|}{\textbf{\begin{tabular}[c]{@{}c@{}}Correspondence to \\      Full Macaque\end{tabular}}} &
  \multicolumn{1}{c}{\textbf{Horse 10}} &
  \multicolumn{1}{c}{\textbf{\begin{tabular}[c]{@{}c@{}}Deep Horse \\      (Ours)\end{tabular}}} &
  \multicolumn{1}{c}{\textbf{\begin{tabular}[c]{@{}c@{}}Correspondence to \\      Horse 10\end{tabular}}} \\ \hline
Nose           & Nose              & Nose           & Nose             & Nose            & Nose             \\
Left Eye       & Neck              &                & Eye              & Head            & Head             \\
Right Eye      & Left Scapula      & Left Shoulder  & Nearknee         & Neck Top        &                  \\
Left Ear       & Right Scapula     & Right Shoulder & Nearfrontfetlock & Neck Middle     &                  \\
Right Ear      & Left Humerus      &                & Nearfrontfoot    & Neck Low        & Mid shoulder      \\
Left Shoulder  & Right Humerus     &                & Offknee          & Right Clavicle  &                  \\
Right Shoulder & Left Forearm      & Left Elbow     & Offfrontfetlock  & Right Upperarm  &                  \\
Left Elbow     & Right Forearm     & Right Elbow    & Offfrontfoot     & Right Forearm   & Nearknee         \\
Right Elbow    & Left Hand         & Left Wrist     & Shoulder         & Right Foreankle & Nearfrontfetlock \\
Left Wrist     & Right Hand        & Right Wrist    & Mid shoulder      & Right Forefeet  & Nearfrontfoot    \\
Right Wrist    & Spine Top         &                & Elbow            & Left Clavicle   &                  \\
Left Hip       & Spine Middle      &                & Girth            & Left Upperarm   &                  \\
Right Hip      & Spine Bottom      &                & Wither           & Left Forearm    & Offknee          \\
Left Knee      & Left Thigh        & Left Hip       & Nearhindhock     & Left Foreankle  & Offfrontfetlock  \\
Right Knee     & Right Thigh       & Right Hip      & Nearhindfetlock  & Left Forefeet   & Offfrontfoot     \\
Left Ankle     & Left Knee         & Left Knee      & Nearhindfoot     & Spine Top       &                  \\
Right Ankle    & Right Knee        & Right Knee     & Hip              & Spine Middle    &                  \\
               & Left Ankle        & Left Ankle     & Stifle           & Spine End       &                  \\
               & Right Ankle       & Right Ankle    & Offhindhock      & Penvis          & Ischium          \\
               & Right Foot        &                & Offhindfetlock   & Right Thigh     &                  \\
               & Left Foot         &                & Offhindfoot      & Right Calf      &                  \\
               & Pelvis            &                & Ischium          & Right Backarm   & Nearhindhock     \\
               & Tail Top          &                &                  & Right Backankle & Nearhindfetlock  \\
               & Tail Upper        &                &                  & Right Backfeet  & Nearhindfoot     \\
               & Tail Upper Middle &                &                  & Left Thigh      &                  \\
               & Tail Middle       &                &                  & Left Calf       &                  \\
               & Tail Lower Middle &                &                  & Left Backarm    & Offhindhock      \\
               & Tail Lower        &                &                  & Left Backankle  & Offhindfetlock   \\
               & Tail End          &                &                  & Left Backfeet   & Offhindfoot      \\
               &                   &                &                  & Tail Top        &                  \\
               &                   &                &                  & Tail Middle     &                  \\
               &                   &                &                  & Tail Low        &                  \\
               &                   &                &                  & Tail End        &                  \\ \hline
\end{tabular}%
}
\vspace{-0.5cm}
\end{table}

\section{Methodology}
The process flow depicted in Figure~\ref{fig:flow} provides an overview of the underlying methodology. In this section, we describe each step in detail and the rationale behind them.

\noindent \textbf{Keypoint Selection.}
In traditional human pose estimation models, a standard set of landmark keypoints is typically used to capture the human skeletal structure effectively \cite{li2019crowdpose,lin2014microsoft}, typical values being 17 for primates and 22 for animals. 
However, these standard anatomical locations do not suffice for pose retargeting on avatar models.
Several animal datasets, such as \cite{labuguen2021macaquepose, yang2022apt}, have significant gaps, particularly in the representation of the spine and tail regions, which are crucial for accurate pose reconstruction in three-dimensional space.
To this end, we have developed a novel set of anatomically important keypoints in a synthetic environment. This expanded keypoint set provides a more comprehensive representation of the primate's full pose, especially with additional keypoints for the spine and tail. 
A comparative analysis with MacaquePose Dataset \cite{labuguen2021macaquepose}, Horse-10 \cite{mathis2021pretraining} and our proposed synthetic dataset is presented in Table \ref{tab:data-corr}. We observe that only a subset of the detected keypoints in the commonly used keypoint system align with anatomical features prevalent in graphics assets. We represent this subset of keypoints as Soft Pose keypoints $\mathcal{K} \in \mathbb{R}^{k_s \times 2}$. 
To address this, we manually add anatomically significant keypoints onto rigged models to prepare synthetic data for transferring poses to 3D avatars. We define this expanded keypoint set $\mathcal{K} \in \mathbb{R}^{k_d \times 2}$ as a Deep Pose Keypoint set. As presented in Table \ref{tab:data-corr}, $(k_s,k_d) = (13,29)$ and $(k_s,k_d) = (16,33)$ for macaque and horse, respectively. This enhanced keypoint configuration improves pose transfer accuracy and supports precise 3D pose reconstruction. 

\begin{figure*}[t]
\centering
\includegraphics[width=0.85\textwidth]{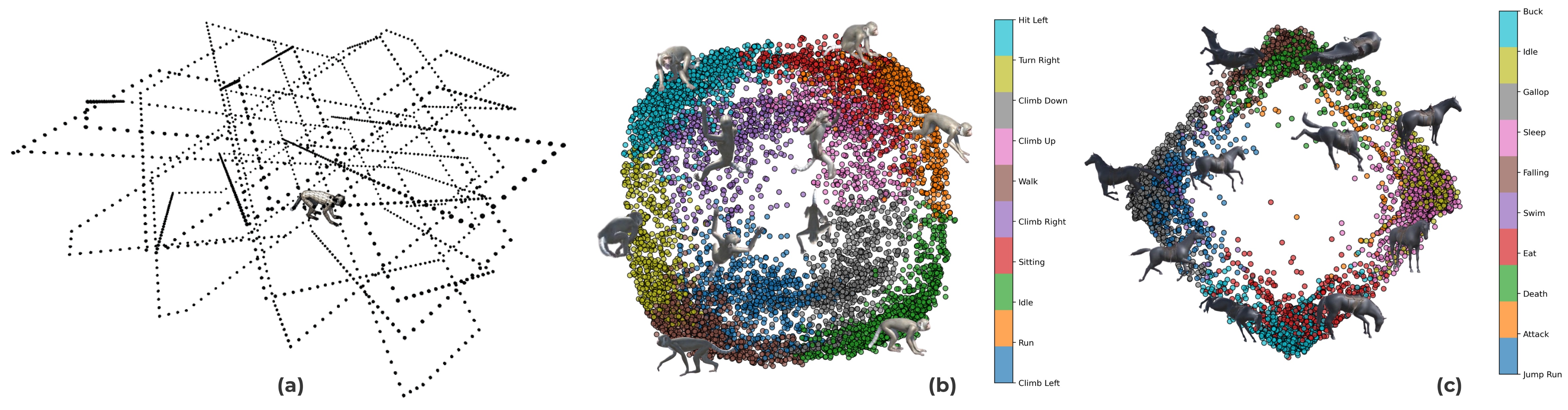}
\caption{ (a) We illustrate our data acquisition setup in Unity, where the subject is positioned at the origin, and the black markers trace the camera's trajectory during data collection. The camera consistently faces the subject, with the field of view calibrated to keep the subject fully visible at all times. (b) and (c) display $k=10$ clusters of 3D poses derived from the look-up table generated from synthetic Macaque and Horse models. In each case, over 60\% of the instances in each cluster correspond to a specific action set. A sample image from each cluster's most frequent action sequence is shown. The scatter points represent the values of the two most dominant principal components obtained from PCA on the look-up table, with colors indicating the cluster to which each scatter point belongs. The corresponding color bar shows the most prominent action in the cluster. Zooming in is recommended for better clarity.
}
\label{fig:cam_knn}
\vspace{-0.3cm}
\end{figure*}

\noindent \textbf{Dataset Construction.}
We begin by defining animation sequences on our avatars that mimic naturalistic actions performed by animals in the wild, specifically focusing on Macaques and Horses for this work. We position the subject at the origin within the 3D environment to ensure accurate data collection and then follow a predetermined camera trajectory, capturing frames at a frequency of 10 frames per second.
We conduct this process across all defined animation sequences to generate distinct poses, described in Table~\ref{tab:pose_details}. This results in a comprehensive dataset comprising 8,000 instances of Macaques and 6,000 instances of Horses. For realistic rendering and simulating physics-aware animation sequences, we utilize Unity Game Engine. Figure~\ref{fig:cam_knn}(a) depicts the camera trajectory followed to generate renderings. 
As we follow this trajectory, we randomly adjust environmental settings, the scale of the subject, and the camera perspectives to introduce diverse variations. In addition to capturing images $\mathcal{I} \in \mathbb{R}^{H \times W}$, we save the previously discussed manually selected keypoints onto the camera's image space to create a set of 2D keypoints $\mathcal{K}^I_{2D} \in \mathbb{R}^{k_s \times 2}$ for each frame. Here, $k_s$ is the number of selected keypoints defined by the Soft Pose keypoint set.
Since image sizes can vary and objects can appear anywhere within the image space, we normalize the coordinates and obtain $\mathcal{K}^n_{2D}$ so that their range extends from [0,1], based on the maximum extent among detected keypoints. The counterpart 3D pose data in world coordinates, denoted as $\mathcal{K}^n_{3D} \in \mathbb{R}^{k_d \times 3}$ is normalized within a unit cube to maintain scale and positional representation uniformity. 
We store the normalized 3D coordinates in a Look-up Table to assist in inferring deep 3D poses from soft 3D poses (discussed subsequently), as illustrated in Figure \ref{fig:flow}(b)

\noindent \textbf{Lifting to 3D Pose using Attention MLP.}
As illustrated in Figure \ref{fig:flow}(a), the core of our methodology involves the use of a modified Attention-based Multi-Layer Perceptron (Attention MLP) network, which facilitates the lifting of 2D keypoints $(x, y)$ coordinates in image space to a normalized 3D space. 
We train Attention MLP on training split of synthetic data utilizing soft-keypoint set $\mathcal{K}^n_{2D} \in \mathbb{R}^{k_S \times 2}$ as input and predict an estimate of 3D pose represented by $\hat{\mathcal{K}}^n_{3D} \in \mathbb{R}^{k_s \times 3}$. We train our network by employing mean square error (MSE) loss between prediction $\hat{\mathcal{K}}^n_{3D} \in \mathbb{R}^{k_s \times 3}$ and corresponding synthetically generated ground-truth $\mathcal{K}^n_{3D} \in \mathbb{R}^{k_s \times 3}$.
It is important to note that the training of our network is image-independent, and learning relies solely on the geometric relationships inherent in the anatomical structure of the subjects. 
As presented in Table \ref{tab:data-corr}, we impose a loss on key points belonging to the soft keypoint set, i.e., $k_s$ out of $k_d$ while lifting to 3D. 
To derive the deep pose necessary for retargeting pose on avatars from the lifted soft pose keypoints, we utilize a pre-generated lookup table, which we explain in detail below.

\noindent \textbf{Dataset Look-up and Keypoint Matching.}
During inference, given a frame from natural image $\mathcal{I} \in \mathbb{R}^{H \times W}$, we apply 2D pose estimation techniques, specifically DeeplabCut \cite{mathis2018deeplabcut}, pretrained on the MacaquePose dataset, and the ViT-Pose model \cite{xu2023vitposepp}, pre-trained on APT-36k \cite{yang2022apt}, to detect 2D keypoints, $\mathcal{K}^I_{2D} \in \mathbb{R}^{k_s \times 2}$, in the image space, and normalize them to the range $[0,1]$, resulting in $\mathcal{K}^n_{2D}$. 
We then extract the $k_s$ keypoints relevant to our task from this set and project them into 3D space using our Attention MLP network to obtain Soft 3D Pose. 
We perform a nearest-neighbor search from values in our look-up table to find the closest match of the predicted soft 3D pose. This look-up process ensures that the selected set accurately transfers on avatars from an input image. This is illustrated in Figure \ref{fig:flow}(b).

\begin{table}[t]
\centering
\caption{Pose-wise average variance in keypoint coordinates across Datasets on animation sequences defined on rigged avatar}
\label{tab:pose_details}
\resizebox{0.9\columnwidth}{!}{%
\begin{tabular}{cccc|cccl}
\hline
\multicolumn{4}{c|}{\textbf{Deep   Macaque}}                       & \multicolumn{4}{c}{\textbf{Deep Horse}}                                                            \\ \hline
\multicolumn{1}{c|}{\textbf{Pose}} &
  \textbf{$\sigma_{x}$} &
  \textbf{$\sigma_{y}$} &
  \textbf{$\sigma_{z}$} &
  \multicolumn{1}{c|}{\textbf{Pose}} &
  \textbf{$\sigma_{x}$} &
  \textbf{$\sigma_{y}$} &
  \multicolumn{1}{c}{\textbf{$\sigma_{z}$}} \\ \hline
\multicolumn{1}{c|}{Attack}       & 0.084 & 0.042 & 0.074 & \multicolumn{1}{c|}{Attack}   & 0.097 & 0.014 & 0.098 \\
\multicolumn{1}{c|}{Climb Down}   & 0.066 & 0.019 & 0.056 & \multicolumn{1}{c|}{Buck}     & 0.089 & 0.019 & 0.090 \\
\multicolumn{1}{c|}{Climb Right}  & 0.077 & 0.019 & 0.060 & \multicolumn{1}{c|}{Death}    & 0.091 & 0.048 & 0.091 \\
\multicolumn{1}{c|}{Climb Up}     & 0.067 & 0.018 & 0.053 & \multicolumn{1}{c|}{Eat}      & 0.097 & 0.016 & 0.096 \\
\multicolumn{1}{c|}{Climg Left}   & 0.076 & 0.021 & 0.063 & \multicolumn{1}{c|}{Falling}  & 0.084 & 0.060 & 0.084 \\
\multicolumn{1}{c|}{Hit Left}     & 0.089 & 0.036 & 0.075 & \multicolumn{1}{c|}{Gallop}   & 0.085 & 0.008 & 0.083 \\
\multicolumn{1}{c|}{Hit Right}    & 0.089 & 0.037 & 0.076 & \multicolumn{1}{c|}{Idle}     & 0.101 & 0.008 & 0.097 \\
\multicolumn{1}{c|}{Idle}         & 0.098 & 0.034 & 0.076 & \multicolumn{1}{c|}{Jump}     & 0.094 & 0.027 & 0.093 \\
\multicolumn{1}{c|}{Jump Forward} & 0.088 & 0.045 & 0.078 & \multicolumn{1}{c|}{Jump Run} & 0.088 & 0.020 & 0.087 \\
\multicolumn{1}{c|}{Jump Inplace} & 0.072 & 0.038 & 0.064 & \multicolumn{1}{c|}{Sleep}    & 0.102 & 0.005 & 0.098 \\
\multicolumn{1}{c|}{Jump Run}     & 0.088 & 0.045 & 0.079 & \multicolumn{1}{c|}{Swim}     & 0.090 & 0.006 & 0.083 \\
\multicolumn{1}{c|}{Run}          & 0.087 & 0.042 & 0.074 & \multicolumn{1}{c|}{Walk}     & 0.103 & 0.005 & 0.096 \\
\multicolumn{1}{c|}{Sitting} &
  0.089 &
  0.029 &
  0.070 &
  \multicolumn{1}{c|}{} &
  \multicolumn{1}{l}{} &
  \multicolumn{1}{l}{} &
  \multicolumn{1}{l}{} \\
\multicolumn{1}{c|}{Turn Left}    & 0.087 & 0.045 & 0.075 & \multicolumn{1}{c|}{}         &       &       &       \\
\multicolumn{1}{c|}{Turn Right}   & 0.088 & 0.044 & 0.076 & \multicolumn{1}{c|}{}         &       &       &       \\
\multicolumn{1}{c|}{Walk}         & 0.083 & 0.037 & 0.072 & \multicolumn{1}{c|}{}         &       &       &       \\ \hline
\end{tabular}%
}
\vspace{-0.4cm}
\end{table}

\noindent \textbf{Pose and Texture Retargeting.}
Finally, we calculate the angles between each joint in the lifted 3D pose. Using the detailed 3D keypoints and the computed joint angles, we implement a pose transfer mechanism in Unity to ensure that the joint angles of our rigged model match those of the lifted pose. 

Deep 3D pose reveals the animal's 3D posture, camera angle, and scale difference between the real and virtual worlds. Texture appearance is an interaction between the physical structure of the surface (e.g., fur), color pigmentation, and the interplay between the illumination source and 3D surfaces. Texture transfer requires 2D-3D corresopndence and knowledge of the animal's 3D posture, camera angle, and scale difference between the real and virtual worlds, these are revealed by Deep 3D pose. We used the avatar’s UV map as a canonical representation for variations in skin structure and used histogram equalization to transfer color tones from the natural image to the avatar. Finally, the rendering process in Unity ensures correct variations in shadows around the avatar’s body.

\begin{table}[]
\centering
\caption{Performance of Different Models on Deep Macaque and Deep Horse Datasets}
\label{tab:sota}
\resizebox{0.9\columnwidth}{!}{%
\begin{tabular}{c|ccc|ccc}
\hline
                    & \multicolumn{3}{c|}{\textbf{Deep Macaque}}         & \multicolumn{3}{c}{\textbf{Deep Horse}}            \\ \cline{2-7} 
\textbf{Model}      & \textbf{MSE}   & \textbf{PDJ@0.2} & \textbf{PDJ@0.05} & \textbf{MSE}   & \textbf{PDJ@0.2} & \textbf{PDJ@0.05} \\ \hline
\textbf{Julieta et. al. \cite{martinez2017simple}} & 0.021 & 0.856 & 0.376  & 0.064 & 0.767 & 0.355 \\
\textbf{GCN \cite{kipf2016semi}}       & 0.028 & 0.791 & 0.242 & 0.062 & 0.645 & 0.311 \\
\textbf{Chang et. al. \cite{chang20182d}}            & 0.020 & 0.867 & 0.391 & 0.060 & 0.791 & 0.370 \\
\textbf{Ours (w/o Att.)} & 0.028 & 0.808 & 0.321  & 0.075 & 0.732 & 0.336 \\
\textbf{Ours (H=2)}     & \textbf{0.015} & 0.892 & 0.394  & 0.040 & 0.733 & 0.249 \\
\textbf{Ours (H=4)} &  0.016 & \textbf{0.899}  & \textbf{0.439}  & \textbf{0.039} & \textbf{0.832}  & \textbf{0.448}  \\ \hline
\end{tabular}%
}
\vspace{-0.4cm}
\end{table}

\section{Results and Discussion}
\noindent \textbf{Efficacy of Lookup Table.}
Although the proposed simple lightweight Attention MLP network can lift 2D keypoints to 3D efficiently, to transfer pose on avatars, we need dense locations of 2D-keypoints that are not detected by existing pose estimation methods.
Since anatomical constraints limit the kinematic variability of joints in animals and humans, we leverage this fact to generate a large lookup table of poses. In Figure \ref{fig:cam_knn}(b) and Figure \ref{fig:cam_knn}(c), we illustrate how poses in the lookup tables are effectively clustered, with similar poses forming distinct clusters. We present a visualization of such a cluster in 2D space by plotting the two most dominant principal components of the lookup table. 
The clustering of similar poses and the distinct separation between clusters help us effectively map soft poses to accurate deep poses using the lookup table. We present qualitative results obtained after each step in Figure \ref{fig:qual}.

\noindent \textbf{Quantitative Comparisons.}
The 2D keypoint coordinate pose data from the Deep Macaque and Horse datasets is divided into a training set (80\% of the data) and a validation set (20\% of the data). For performance evaluation, we specifically use methods that, like ours, focus on lifting 2D keypoint sets to 3D poses and do not rely on image data.
We first train the networks proposed by \cite{martinez2017simple, chang20182d}, along with baseline models, on our Deep Macaque and Horse datasets on the training splits. 
All variants in the table are trained for 100 Epochs.
We present quantitative comparisons in terms of Mean Squared Error (MSE) and Percentage of Detected Joint (PDJ) with the proposed Attention MLP network in Table \ref{tab:sota}.
Here, PDJ@x signifies a correct keypoint detection if it lies within x × d distance, where d represents the diagonal of the bounding box containing the target and $H$ is the number of heads used in MHSA Block.
Methods that directly regress 3D keypoints from input images, such as \cite{sun2018integral, zhou2019hemlets, jiang2023rtmpose}, are generally very resource-intensive and do not match the context of the pose lifting problem, so we do not include them in our comparisons. 
In addition to comparisons with state-of-the-art methods, Table \ref{tab:sota} also provides quantitative results from an ablation study of various baseline networks. 
We find that the MLP integrated with the attention module outperforms existing approaches.

% \begin{enumerate} 
%     \item We denote a hidden layer with A input channels and B output channels with a dropout rate of 0.2 as A-B. Table~\ref{tab:performance} provides a quantitative comparison of the results across different model architectures. The MLPh1 model is a simple MLP with a single hidden layer of size 512-512 between the input and output layers. MLPh2 consists of two hidden layers, each with 512-512 units, while MLPh3 incorporates three such hidden layers. Similarly, MLPh4 employs four 512-512 hidden layers between the input and output layers. 
% \end{enumerate}

\section{Conclusion}
This work introduced a method for seamlessly retargeting poses from a natural image to 3D virtual avatars. To overcome the challenges of acquiring 3D annotated data for animals, we developed a pipeline to generate synthetic data and develop two datasets, Deep Macaque and Deep Horse (together L3D-Pose dataset), using a variety of action sequences defined on rigged models. Our approach employs a lookup table to map soft poses to deep poses, effectively transferring poses from natural images to synthetic avatars. Additionally, our method lifts 2D keypoints to 3D poses independently of the input image, providing better scalability in scenarios with limited data. Because our approach does not rely on images, it effectively generalizes from training on synthetic data to delivering accurate inferences on the natural images captured in the wild\footnote{\footnotesize This work is supported by the Jibaben Patel Chair in AI held by Shanmuganathan Raman.}.

\bibliographystyle{IEEEtran}
\bibliography{refs}

\end{document}